\documentclass[sigconf,authorversion]{acmart}

\usepackage{graphicx}
\usepackage{booktabs}
\usepackage{multirow}
\usepackage{multicol}
\usepackage{adjustbox}
\usepackage{url}
\usepackage{balance}
\usepackage{amsmath}
\usepackage{xspace}
\usepackage{enumitem}
\usepackage{csquotes}
\AtBeginDocument{%
  }

\copyrightyear{2024}
\acmYear{2024}
\setcopyright{acmlicensed}
\acmConference[CIKM '24] {Proceedings of the 33rd ACM International Conference on Information and Knowledge Management}{October 21--25, 2024}{Boise, ID, USA.}
\acmBooktitle{Proceedings of the 33rd ACM International Conference on Information and Knowledge Management (CIKM '24), October 21--25, 2024, Boise, ID, USA}
\acmISBN{979-8-4007-0436-9/24/10}
\acmDOI{10.1145/3627673.3679894}
\acmSubmissionID{sp2631}

\begin{document}

\title{CXSimulator: A User Behavior Simulation using LLM Embeddings for Web-Marketing Campaign Assessment}

\author{Akira Kasuga}
\orcid{0009-0003-8870-8010}
\affiliation{%
  \institution{CyberAgent, Inc.}
  \city{Tokyo}
  \country{Japan}
}
\email{kasuga_akira@cyberagent.co.jp}

\author{Ryo Yonetani}
\orcid{0000-0002-2724-6233}
\affiliation{%
  \institution{CyberAgent, Inc.}
  \city{Tokyo}
  \country{Japan}
}
\email{yonetani_ryo@cyberagent.co.jp}

\renewcommand{\shortauthors}{Kasuga and Yonetani}

\def\eg{{\it e.g.}}
\def\cf{{\it c.f.}}
\def\ie{{\it i.e.}}
\def\etal{{\it et al. }}
\def\etc{{\it etc}}

\begin{abstract}
This paper presents the Customer Experience (CX) Simulator, a novel framework designed to assess the effects of untested web-marketing campaigns through user behavior simulations. The proposed framework leverages large language models (LLMs) to represent various events in a user's behavioral history, such as viewing an item, applying a coupon, or purchasing an item, as semantic embedding vectors. We train a model to predict transitions between events from their LLM embeddings, which can even generalize to unseen events by learning from diverse training data. In web-marketing applications, we leverage this transition prediction model to simulate how users might react differently when new campaigns or products are presented to them. This allows us to eliminate the need for costly online testing and enhance the marketers' abilities to reveal insights. Our numerical evaluation and user study, utilizing BigQuery Public Datasets from the Google Merchandise Store, demonstrate the effectiveness of our framework. 
\end{abstract}



\maketitle
\section{Introduction}
\label{sec:intro}

Sophisticated web marketing tools, such as Google Analytics~\cite{GA} and Braze~\cite{Braze}, have been used to measure the behavior of users and assess marketing strategies. However, only fully developed web services can use these tools optimally. A/B testing, though commonly employed, is expensive to execute repeatedly and challenging to ensure a sufficient number of target users. Expert marketing knowledge and experience are necessary to draw useful insights from user behavior. As a consequence, it is not trivial to benefit from such tools in the early stages of service or in situations where there is a shortage of experts.

We focus on the potential of LLMs to solve this issue. LLMs have been applied not only for natural language processing tasks~\cite{hadi2023survey} but also for common sense reasoning in multi-modal data~\cite{wang2024exploring}. We believe that the ability of LLMs, especially to represent the high-level semantics of complex event descriptions with compact embedded vectors (\ie, LLM embeddings)~\cite{keraghel2024beyond}, can also be advantageous for web marketing applications

\begin{figure*}[t]
  \centering
  \includegraphics[width=\linewidth]{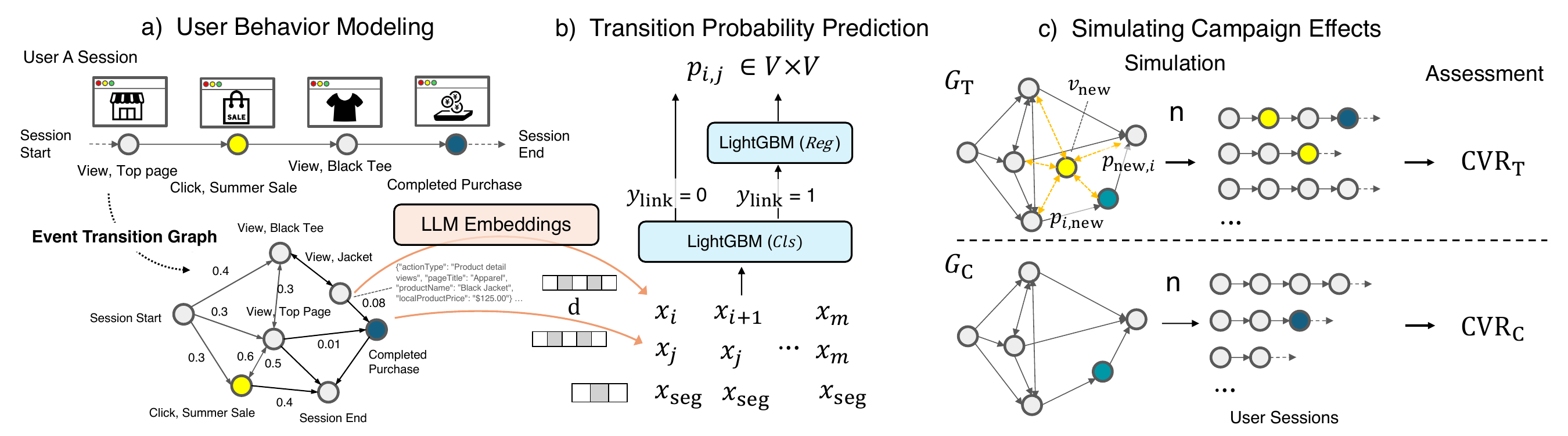}
  \caption{CXSimulator Framework: a) Constructing an event transition graph, b) Hybrid model for regressing all transition probabilities, c) Simulating user behaviors and assessing a campaign with a control and a treatment group.}
  \label{fig:framework}
\end{figure*}

In this work, we propose a novel framework named \textbf{CXSimulator} (Figure~\ref{fig:framework}). CXSimulator can facilitate a fast offline assessment of the effects of marketing campaigns via LLM-based user behavior simulation, in the absence of costly A/B testing and expert knowledge. The proposed framework describes user behaviors on e-commerce sites with an event transition graph, where nodes represent individual events such as viewing an item and clicking a coupon, while edges indicate the transition from one event to another. In the graph, we embed the detailed description of each event into a semantic latent vector using LLMs, and train transition prediction models that use the LLM embeddings to predict connections and transition probabilities between events. The trained models can be employed to simulate plausible user behaviors, even when the graph includes a new node absent in the training data. This way, our approach can assess the effectiveness of new, untested campaigns by comparing simulated user behaviors with and without the nodes and measuring the difference in conversion rates.

We evaluate the effectiveness of the proposed CXSimulator using BigQuery Public Datasets~\cite{BigQueryPublicDatasets} from GA360~\cite{BigQueryGA}, which comprises real user behavior histories on the Google Merchandise Store. The experimental results reveal that our transition prediction models outperform relevant link prediction methods~\cite{grover2016node2vec,pan2018adversarially,li2024condensed}, as well as other LLM-based solutions using GPT-3.5/4~\cite{AzureOpenAIGPT3.5,AzureOpenAIGPT4}. Furthermore, our user study with five domain experts validates a high correlation between the CXSimulator's assessments and expert marketers' judgments regarding untested campaign effectiveness. The contributions of this paper are summarized as follows:

\begin{itemize}
\item We present CXSimulator, a framework that models user behaviors on e-commerce sites using event transition graphs.
\item We demonstrate the effectiveness of using LLM embeddings to predict transition probabilities between events, which allows us to simulate user behaviors even in the presence of a new and untested campaign.
\item We evaluate the effectiveness of a new campaign between the proposed method and domain experts.
\end{itemize}

\section{Related Work}
\label{sec:related_work}

\noindent\textbf{User Behavior Modeling and Simulation.}
The recent growth in e-commerce has highlighted the importance of analyzing extensive user behavior logs~\cite{E-commerce}. User behavior modeling is employed in search~\cite{ZHANG202040} and recommendation systems~\cite{UBMforRec}. Recent work has investigated methods that employ graph neural networks~\cite{wu2019session} and long short-term memory~\cite{panagiotakis2022session} to session-based user behavior histories. Moreover, LLM technologies have been advantageous for web content generation~\cite{NLPForMarketing} and creative image generation~\cite{onlineAdsLLM}. However, online evaluation methods present challenges due to their high costs and scalability issues~\cite{kohavi2020trustworthy}. Conversely, there has recently been increasing interest in offline evaluation through simulation. Simulators have been used for evaluating conversational recommendation systems~\cite{afzali2023usersimcrs} and information retrieval methods~\cite{SimIAS2023}. Despite such developments, research has been limited in scope. This background motivates us to employ marketing campaign simulation to address this research gap.

\noindent\textbf{Link Prediction,} which aims to anticipate relationships between nodes, serves various purposes such as network reconstruction, recommendation systems, and knowledge graph completion~\cite{kumar2020link}. In e-commerce applications, the primary focus is on predicting purchase and viewing relationships between users and items~\cite{li2009recommendation}. Nevertheless, studies on predicting transition probabilities for behavior transition graphs are scarce. Existing methods use Node2Vec~\cite{grover2016node2vec} or graph auto-encoder~\cite{pan2018adversarially} for link prediction~\cite{trouillon2016complex}, but these approaches have limitations in representing knowledge outside of the training data. Recently, methods have been developed to directly predict links by fine-tuning node features and graph structures as prompts in knowledge graphs~\cite{shu2024knowledge}, as well as techniques for direct link prediction in zero-shot scenarios~\cite{li2024condensed}. Consequently, this paper leverages the potential of LLMs for common sense reasoning~\cite{chen2024exploring}, and employs LLM embeddings of user behavior for link prediction.
\section{CXSimulator Framework}
\label{sec:proposed}

As illustrated in Figure~\ref{fig:framework}, the CXSimulator incorporates three main components: a) user behavior modeling based on an event transition graph, b) prediction of transition probabilities through LLM embeddings, and c) user behavior simulation for marketing campaign assessment. Each of these components is detailed below.

\subsection{User Behavior Modeling}
Consider users who are interacting with an e-commerce service. Their behavior history is represented as a sequence of events such as \emph{``viewing page A'', ``obtaining coupon B'', ``adding item C to the cart'', and ``checking out the item''}. A weighted directed graph naturally serves to represent the transitions among these events. Specifically, let $G=(V, E)$ be a digraph, referred to as the \emph{event transition graph}, where $V$ denotes the nodes each corresponding to individual events. $E=V \times V$ indicates the edges that stand for transitions between events. Each edge carries a transition probability, $p_{i, j} \in (0, 1]$ for a pair of vertices $v_i, v_j \in V, \{i,j \mid 1,2,\ldots,\mathrm{m}\}$. The existence of edges and probabilities for linked nodes can be acquired from a corpus of user behavior data in the same segment such as cities and devices. Once this is achieved, a plausible user behavior history can be simulated by traversing the event transition graph according to these transition probabilities.

\subsection{Transition Probability Prediction}
Our key question arises here: \emph{how can we simulate user behaviors if we add a new event to the event transition graph?} Such situations would frequently occur in web-marketing scenarios, for example, when a new item is put on sale or a new discount campaign is launched. The effects of events, \ie, whether these events contribute to the conversion rate, can vary based on user behaviors within a given segment. Ideally, we expect to simulate such effects \emph{before} the items/campaigns are presented to real users. 

This situation can be viewed as a variant of link prediction problems~\cite{kumar2020link}, where appropriate feature extraction for nodes is essential for high prediction performance. In this work, we propose to leverage the reasoning ability of LLMs to represent the semantic aspects of events and user segments with compact embeddings. Google Analytics provides detailed events in a well-structured format, such as \texttt{\small{\{"actionType": "Product detail views", "pageTitle": "Office", "pagePath": "/store.html/quickview" , "productName": "Colored Pencil Set", "localProductPrice": "\$3.99"\}}}. Likewise, some attributes of user segments are readily available in a similar format, such as \texttt{\small{\{"country": "United States", "browser": "Chrome", "source": "Direct"\}}}. We transform such events and segment descriptions into LLM embeddings, and employ them as node features. 

Formally, we solve two-fold tasks: 1) binary classification for the existence of edges between a pair of nodes, and 2) regression for the transition probabilities of the edges when they exist. Let $f_i, f_j, f_\mathrm{seg} \in F$ be the LLM embedding vectors from nodes $v_i, v_j$ as well as the segment attribute, where $F\in\mathbb{R}^\mathrm{d}$ is the d-dimensional embedding space. We concatenate these vectors to use as the input to classifier $\mathit{Cls}:F\times F\times F \rightarrow \{0, 1\}$ and regressor $\mathit{Reg}:F\times F\times F \rightarrow [0, 1]$. Note that filtering the input with the classifier before regressing the transition probabilities yields better performance than directly performing the regression for all possible pairs of nodes, regardless of the existence of edges. This is due to the wide variety of nodes, the sparsity of edges, and the resulting data imbalance.

\subsection{Simulating Campaign Effects}
This section describes how to simulate user behavior for a new event and how to utilize these simulations to assess web-marketing campaigns. To achieve this, we consider two event transition graphs: the Control $G_\mathrm{C}=(V_\mathrm{C}, E_\mathrm{C})$, where the edge connectivity and weights are derived solely from user behavior histories, and the Treatment $G_\mathrm{T}=(V_\mathrm{T}, E_\mathrm{T})$, which has an expanded node and edge collection with a new event $v_\mathrm{new}$. While it is simply $V_\mathrm{T}=V_\mathrm{C}+\{v_\mathrm{new}\}$, we span edges between $v_i\in V_\mathrm{C}$ and  $v_\mathrm{new}$ based on the output from the trained models of $\mathit{Cls}(f_i, f_\mathrm{new}, f_\mathrm{seg})$ and $\mathit{Cls}(f_\mathrm{new}, f_i, f_\mathrm{seg})$, and also estimate the transition probability with the trained $\mathit{Reg}(f_i, f_\mathrm{new}, f_\mathrm{seg})$ and $\mathit{Reg}(f_\mathrm{new}, f_i, f_\mathrm{seg})$ for the spanned bidirectional edges, where $f_\mathrm{new}$ is the LLM embedding vector on $v_\mathrm{new}$.

Beginning from the same event node (\eg, \emph{``session start''}), we sample multiple event sequences from both $G_\mathrm{C}$ and $G_\mathrm{T}$ until we reach the ending node (\eg, \emph{``session end''}) or exceed the maximum sequence length. Each sequence signifies plausible user behavior with or without the intervention of the new event $v_\mathrm{new}$. A collection of these sequences allows us to predict the potential difference in conversion rates using standard marketing analysis methods.
\section{Experiment}
\label{sec:experiment}

We employed the BigQuery Public Datasets~\cite{BigQueryPublicDatasets} from GA360~\cite{BigQueryGA} to conduct numerical experiments on transition probability prediction and a user study for assessment of new campaigns.

\subsection{Numerical Evaluation}
\textbf{Datasets.}
The dataset~\cite{BigQueryGA} includes e-commerce events (\eg, click through of product lists, product detail views, add product(s) to cart, check out, and completed purchase) and detailed descriptions (\eg, pageTitle, pagePath, productName, localProductPrice,\ldots). It spans a total period of 366 days, from 2016/08/01 to 2017/08/01, amounting to 4,153,675 records. We designated the training, validation, and testing periods as 2016/08, 2016/09, and 2016/10, respectively, ensuring no seasonal differences in the data during these periods.\footnote{The limited granularity of event descriptions (due to ID mapping) prevented the utilization of other popular datasets, including CIKM2016EComm~\cite{CIKM2016EComm,codalab_competitions_JMLR} and REES46~\cite{REES46}.}

\noindent\textbf{Implementation Details.}
We implemented $\mathit{Cls}$ and $\mathit{Reg}$ using LightGBM~\cite{GuolinLightGBM}, a lightweight and fast model to ensure timely execution that preserves user experience. In the training data, the total number of nodes amounted to 1,350, with 19,126 linked edges out of total 1,822,500 edges. We applied Laplace smoothing $\alpha=\mathrm{5}$ to mitigate the zero frequency issue due to data sparsity. The class weight in \emph{Cls} was considered based on edge sparsity and imbalance. The event description was vectorized to \(\mathrm{d}=128\) by Azure OpenAI Service ~\cite{AzureOpenAIService,AzureOpenAIEmbeddingSmall,AzureOpenAIEmbeddingLarge}.
The main parameters are, for $\mathit{Cls}$, \{\enquote{objective}: \enquote{binary},  \enquote{scale\_pos\_weight}: 5.0, \enquote{num\_iteration}: 1500, \enquote{metric}: \enquote{auc}\}, and for $\mathit{Reg}$, \{\enquote{objective}: \enquote{regression}, \enquote{num\_iteration}: 1500, \enquote{metric}: \enquote{rmse}\}. Both models have the \enquote{early\_stopping\_round} parameter set to 50. We validated prediction accuracy for transition probabilities \(p_{i,j}\) of all node pairs \(v_i,v_j\) in the testing period. Furthermore, we compare accuracy using all combinations of \emph{text-embedding-3-small} (v3-small)~\cite{AzureOpenAIEmbeddingSmall} and \emph{text-embedding-3-large} (v3-large)~\cite{AzureOpenAIEmbeddingLarge}, as well as either $\mathit{Cls}$ or $\mathit{Reg}$ solely.

\noindent\textbf{Baselines.}
As baseline methods, we predict all transition probabilities \(p_{i,j}\) using Node2Vec~\cite{grover2016node2vec} based on the graph structure, and graph auto-encoder (GAE)~\cite{pan2018adversarially} that considers both the graph structure and node descriptions. With Node2Vec and GAE, an embedding vector of \(\mathrm{d}=16\) effectively described a node feature. 
Moreover, we adopted GPT-3.5/4~\cite{AzureOpenAIGPT3.5,AzureOpenAIGPT4} via few/zero-shot~\cite{shu2024knowledge,li2024condensed} with prompts\footnote{The prompt begins as follows: \enquote{Given a src node description and dst node description, your task is to predict transition probability(0.0-1.0) of the edge between 2 nodes.}} consisting of detailed event description to directly predict \(p_{i,j}\).

\noindent\textbf{Evaluation Metrics.}
Our evaluation metrics include: 
\textbf{RMSE} minimizes the error of \(p_{i,j}\), a continuous value. 
\textbf{SMAPE}~\cite{kreinovich2014estimate} evaluates relative error in probability regression \(p_{i,j}\). 
\textbf{F1-Score} represents the harmonic mean of precision and recall. 
\textbf{ROC-AUC} measures a model's ability to correctly classify between positive and negative cases.
\textbf{PR-AUC} provides a more appropriate metric in situations with class imbalance than ROC-AUC.
We adjusted the decision threshold~\cite{esposito2021ghost} in $\mathit{Reg}$ and predicted probability in $\mathit{Cls}$ to align with their objectives and evaluate all metrics consistently across all methods. The optimized decision threshold and predicted probability were also 0.1 in this experiment.

\noindent\textbf{Results.}
Table~\ref{table:prediction-result} demonstrates that our models outperform the baseline methods. Using $\mathit{Cls}$ and $\mathit{Reg}$ with v3-small is deemed more practical due to cost efficiency, since the difference in metrics between v3-small and v3-large is not significant. The GPT-3.5/4 baseline showed limited performances because it could only receive 20-30 samples in a prompt.

\subsection{User Study}
We conducted a user study with marketers to validate the practical utility of the CXSimulator in actual business environments.

\noindent\textbf{Participants.} We recruited five marketers where three are senior (over 3 years of experience) and two are junior ones (1-2 years of experience). They participated in the two studies described below.

\noindent\textbf{Study 1: Validity of simulating the difference in conversion rates.}
In this study, we assess the concordance between the difference in conversion rates simulated using the CXSimulator framework and the judgements of marketers. Each participant was asked to assess the effectiveness of 35 campaigns which have never been implemented in the store, for example, \emph{\enquote{Enjoy 1 month Free of YouTube Premium for YouTube related Product!}}, by analyzing the GA data from 2016/08/01 to 08/31 on BigQuery within 40 minutes. The 35 campaigns consisted of 5 variations each in 7 categories, and comprehensiveness of the assessment is reasonable. Each campaign was scored on a four-point scale: effective (4), somewhat effective (3), not very effective (2), and not effective (1). For when it was difficult to judge, an option of \enquote{unable to judge} was also provided. On the other hand, we simulated 10,000 user sessions from each of $G_\mathrm{C}$ and $G_\mathrm{T}$ during the same period in our framework and measured conversion rates. Finally, we evaluate the correlation between the difference in conversion rates and human assessments.

\begin{figure}[t]
  \centering
  \includegraphics[width=1.0\columnwidth]{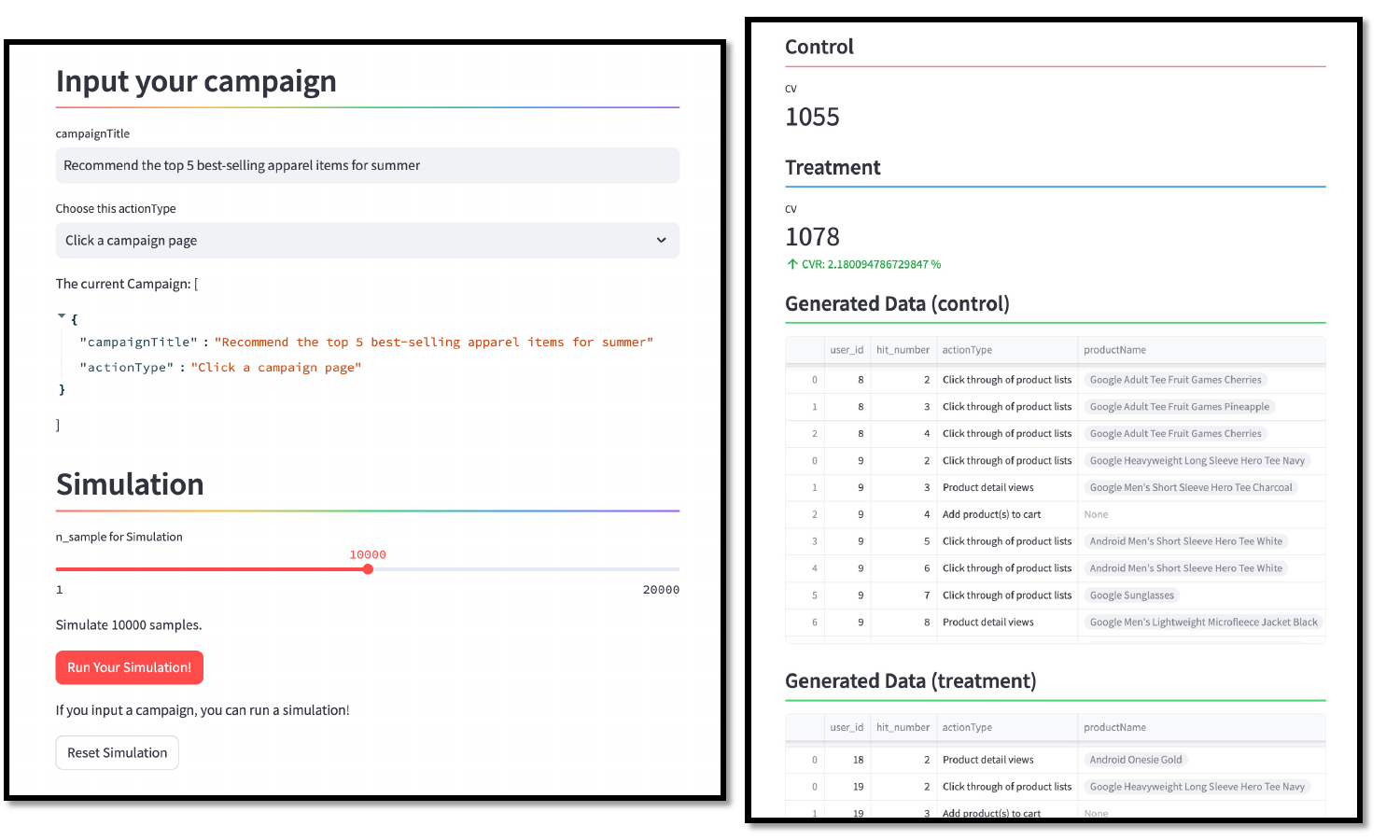}
  \caption{CXSimulator User Interface (UI) : Left) Input your campaigns and run a simulation. Right) Execution results.}
  \label{fig:simulator_ui}
\end{figure}

\noindent\textbf{Study 2: Usefulness of proposed simulator UI.}
Furthermore, we developed a UI to experience the CXSimulator. As shown in Figure~\ref{fig:simulator_ui}, our UI consists of five components: 1) data view and analysis, 2) event input forms (\eg, campaignTitle, actionType), 3) Simulation run button, 4) Control/Treatment CVR comparison, 5) Control/Treatment user behavior data view and analysis.
Participants were asked to input their own campaigns, for example, \emph{\{\enquote{actionType}: Click a campaign page, \enquote{campaignTitle}: Recommend the top 5 best-selling...\}}, and simulate the effect on the UI. After sufficient interactions with the UI, they evaluated planning efficiency (Eff), output data validity (Val), system operability (Ops), function expandability (Ext), and recommendation to another user (Rec) on a 10-point scale from No to Yes within 20 minutes. We also collected responses about the selection rationale through a questionnaire.

\noindent\textbf{Results.}
All the results are summarized in Table~\ref{table:simulation-result}.
The correlation coefficient for evaluations from senior participants was \textbf{0.6220}, while it was \textbf{0.3811} for junior participants. This indicates that the simulator's assessments could be more similar to those of senior marketers. In the questionnaire, the senior participants expressed difficulty in assessing campaigns within a restricted time frame, as they often relied on their intuition rather than conducting a thorough analysis. This reliance on intuition can explain why a higher correlation exists between the seniors and the simulator compared to the juniors. The overall evaluations on the usefulness of our system were also high, pointing to potential contributions to marketing work. Interestingly, junior marketers gave higher ratings to the system than the seniors. These results indicate the potential of our simulator to be a training tool.
\subsection{Limitations and Possible Extensions}
While our system worked reasonably well, we found several failure cases through the user study. For example, the assessments for the campaign \enquote{\emph{Rank up to silver member with an additional \$30 purchase this month!}} were extremely high with 4,3,3,4,4 across all participants, while the simulator outputs a low conversion rate uplift with or without the campaign event.
The observed difficulty might stem from the absence of a real membership system in the store and in the data. This situation is outside the scope of the LLM's common-sense reasoning. Potential extensions to mitigate this problem can be fine-tuning of LLMs to obtain domain-specific contexts.  Moreover, one senior participant rated 4/10 on the validity of developed UI, due to unclear rationale for the conversion rate presented by the simulator. Enhancing interpretability by illustrating variations in statistical trends or extracting significant user behavior in a control and treatment group would be an interesting future direction.

\begin{table}[tbp]
\caption{Numerical Evaluation Results}
\label{table:prediction-result}
\begin{adjustbox}{width=.9\columnwidth}
\centering
\begin{tabular}{@{}lccccc@{}} \toprule
   methods & RMSE $\downarrow$ & SMAPE $\downarrow$ & F1-Score $\uparrow$ & ROC-AUC $\uparrow$ & PR-AUC $\uparrow$ \\ \midrule
   \textbf{Proposed} \\
   v3-small  & & & & & \\
   \ \ \ Cls+Reg & \textbf{0.054} & 23.50 & 0.853 & \textbf{0.937} & 0.810 \\
   v3-large & & & & & \\
    \ \ \ Cls+Reg & 0.055 & \textbf{22.08} & \textbf{0.856} & 0.922 & \textbf{0.828}\\ \midrule
   v3-small & & & & & \\
    \ \ \ Cls & 0.659 & 43.80 & 0.283 & 0.893 & 0.009 \\
    \ \ \ Reg & 0.066 & 34.04 & 0.754 & 0.899 & 0.784 \\
   v3-large & & & & & \\
    \ \ \ Cls & 0.550 & 42.53 & 0.345 & 0.828 & 0.006 \\
    \ \ \ Reg & 0.100 & 35.27 & 0.480 & 0.518 & 0.183 \\
\midrule
   Node2Vec & & & & & \\
    \ \ \ Cls+Reg & 0.085 & 27.00 & 0.582 & 0.750 & 0.388 \\
   GAE & & & & & \\
    \ \ \ Cls+Reg & 0.084 & 32.80 & 0.499 & 0.843 & 0.255\\
    \midrule
   GPT-4 & & & & & \\
    \ \ \ Few-Shot & 0.074 & 23.87 & 0.774 & 0.888 & 0.684 \\
    \ \ \ Zero-Shot & 0.197 & 40.85 & 0.178 & 0.784 & 0.528 \\
   GPT-3.5 & & & & & \\
    \ \ \ Few-Shot & 0.215 & 31.01 & 0.759 & 0.763 & 0.512 \\
    \ \ \ Zero-Shot & 0.543 & 43.49 & 0.149 & 0.663 & 0.456 \\ \bottomrule
 \end{tabular}
 \end{adjustbox}
 \end{table}

\begin{table}[t]
\caption{User Study Results}
\label{table:simulation-result}
\begin{adjustbox}{width=0.85\columnwidth}
\centering
\begin{tabular}{@{}lcccccc@{}} \toprule
   Participants & Correlation & Eff & Val & Ope & Ext & Rec \\ \midrule
   Senior & \textbf{0.6220} & 7.7 & 7.3 & 7.3 & 8.7 & 8.3 \\
   Junior & 0.3811 & \textbf{9.0} & \textbf{7.5} & \textbf{9.5} & \textbf{9.5} & \textbf{9.5} \\ \bottomrule
 \end{tabular}
 \end{adjustbox}
 \end{table}

\section{Conclusion}
\label{sec:concl}

In this work, we propose a novel framework, \textbf{CXSimulator}, designed for offline assessments of untested campaigns through user behavior simulations based on LLM embeddings. 
Our experimental results demonstrate the effectiveness of our transition probability prediction based on LLM embeddings compared to existing link prediction methods. We also confirm a correlation between the assessments from the CXSimulator and those from marketers, thereby indicating the practicability of the proposed framework in real-world marketing operations. In future research, we aim to examine the effectiveness of our approach within the domain and format constraints of private data from actual retail stores.

\balance

\bibliographystyle{ACM-Reference-Format}
\balance
\bibliography{abbrev}


\begin{thebibliography}{36}


\ifx \showCODEN    \undefined \def \showCODEN     #1{\unskip}     \fi
\ifx \showDOI      \undefined \def \showDOI       #1{#1}\fi
\ifx \showISBNx    \undefined \def \showISBNx     #1{\unskip}     \fi
\ifx \showISBNxiii \undefined \def \showISBNxiii  #1{\unskip}     \fi
\ifx \showISSN     \undefined \def \showISSN      #1{\unskip}     \fi
\ifx \showLCCN     \undefined \def \showLCCN      #1{\unskip}     \fi
\ifx \shownote     \undefined \def \shownote      #1{#1}          \fi
\ifx \showarticletitle \undefined \def \showarticletitle #1{#1}   \fi
\ifx \showURL      \undefined \def \showURL       {\relax}        \fi
\providecommand\bibfield[2]{#2}
\providecommand\bibinfo[2]{#2}
\providecommand\natexlab[1]{#1}
\providecommand\showeprint[2][]{arXiv:#2}

\bibitem[Afzali et~al\mbox{.}(2023)]%
        {afzali2023usersimcrs}
\bibfield{author}{\bibinfo{person}{Jafar Afzali}, \bibinfo{person}{Aleksander~Mark Drzewiecki}, \bibinfo{person}{Krisztian Balog}, {and} \bibinfo{person}{Shuo Zhang}.} \bibinfo{year}{2023}\natexlab{}.
\newblock \showarticletitle{Usersimcrs: A user simulation toolkit for evaluating conversational recommender systems}. In \bibinfo{booktitle}{\emph{Proceedings of the 16th ACM International Conference on Web Search and Data Mining}}. \bibinfo{pages}{1160--1163}.
\newblock


\bibitem[Akter and Fosso~Wamba(2016)]%
        {E-commerce}
\bibfield{author}{\bibinfo{person}{Shahriar Akter} {and} \bibinfo{person}{Samuel Fosso~Wamba}.} \bibinfo{year}{2016}\natexlab{}.
\newblock \showarticletitle{Big data analytics in E-commerce: a systematic review and agenda for future research}.
\newblock \bibinfo{journal}{\emph{Electronic Markets}}  \bibinfo{volume}{26} (\bibinfo{year}{2016}), \bibinfo{pages}{173--194}.
\newblock
\urldef\tempurl%
\url{https://doi.org/10.1007/s12525-016-0219-0}
\showDOI{\tempurl}


\bibitem[Balog and Zhai(2023)]%
        {SimIAS2023}
\bibfield{author}{\bibinfo{person}{Krisztian Balog} {and} \bibinfo{person}{ChengXiang Zhai}.} \bibinfo{year}{2023}\natexlab{}.
\newblock \showarticletitle{User Simulation for Evaluating Information Access Systems}. In \bibinfo{booktitle}{\emph{Proceedings of the Annual International ACM SIGIR Conference on Research and Development in Information Retrieval in the Asia Pacific Region}}. \bibinfo{publisher}{Association for Computing Machinery}, \bibinfo{pages}{302--305}.
\newblock
\showISBNx{9798400704086}
\urldef\tempurl%
\url{https://doi.org/10.1145/3624918.3629549}
\showDOI{\tempurl}


\bibitem[Braze, Inc.(2024)]%
        {Braze}
Braze, Inc. \bibinfo{year}{2024}\natexlab{}.
\newblock \bibinfo{booktitle}{\emph{Braze}}.
\newblock Braze, Inc., \url{https://www.braze.com}.
\newblock
\newblock
\shownote{Accessed on April 28, 2024}.


\bibitem[Chen et~al\mbox{.}(2024)]%
        {chen2024exploring}
\bibfield{author}{\bibinfo{person}{Zhikai Chen}, \bibinfo{person}{Haitao Mao}, \bibinfo{person}{Hang Li}, \bibinfo{person}{Wei Jin}, \bibinfo{person}{Hongzhi Wen}, \bibinfo{person}{Xiaochi Wei}, \bibinfo{person}{Shuaiqiang Wang}, \bibinfo{person}{Dawei Yin}, \bibinfo{person}{Wenqi Fan}, \bibinfo{person}{Hui Liu}, {et~al\mbox{.}}} \bibinfo{year}{2024}\natexlab{}.
\newblock \showarticletitle{Exploring the potential of large language models (llms) in learning on graphs}.
\newblock \bibinfo{journal}{\emph{ACM SIGKDD Explorations Newsletter}} \bibinfo{volume}{25}, \bibinfo{number}{2} (\bibinfo{year}{2024}), \bibinfo{pages}{42--61}.
\newblock


\bibitem[Esposito et~al\mbox{.}(2021)]%
        {esposito2021ghost}
\bibfield{author}{\bibinfo{person}{Carmen Esposito}, \bibinfo{person}{Gregory~A Landrum}, \bibinfo{person}{Nadine Schneider}, \bibinfo{person}{Nikolaus Stiefl}, {and} \bibinfo{person}{Sereina Riniker}.} \bibinfo{year}{2021}\natexlab{}.
\newblock \showarticletitle{GHOST: adjusting the decision threshold to handle imbalanced data in machine learning}.
\newblock \bibinfo{journal}{\emph{Journal of Chemical Information and Modeling}} \bibinfo{volume}{61}, \bibinfo{number}{6} (\bibinfo{year}{2021}), \bibinfo{pages}{2623--2640}.
\newblock


\bibitem[Feizi et~al\mbox{.}(2023)]%
        {onlineAdsLLM}
\bibfield{author}{\bibinfo{person}{Soheil Feizi}, \bibinfo{person}{MohammadTaghi Hajiaghayi}, \bibinfo{person}{Keivan Rezaei}, {and} \bibinfo{person}{Suho Shin}.} \bibinfo{year}{2023}\natexlab{}.
\newblock \showarticletitle{Online Advertisements with LLMs: Opportunities and Challenges}.
\newblock \bibinfo{journal}{\emph{arXiv preprint arXiv:2311.07601}} (\bibinfo{year}{2023}).
\newblock


\bibitem[Google LLC(2024a)]%
        {BigQueryPublicDatasets}
Google LLC \bibinfo{year}{2024}\natexlab{a}.
\newblock \bibinfo{booktitle}{\emph{BigQuery public datasets}}.
\newblock Google LLC, \url{https://cloud.google.com/bigquery/public-data}.
\newblock


\bibitem[Google LLC(2024b)]%
        {GA}
Google LLC \bibinfo{year}{2024}\natexlab{b}.
\newblock \bibinfo{booktitle}{\emph{Google Analytics}}.
\newblock Google LLC, \url{https://developers.google.com/analytics}.
\newblock
\newblock
\shownote{Accessed on April 20, 2024}.


\bibitem[Google LLC(2024c)]%
        {BigQueryGA}
Google LLC \bibinfo{year}{2024}\natexlab{c}.
\newblock \bibinfo{booktitle}{\emph{Google Analytics sample dataset for BigQuery}}.
\newblock Google LLC, \url{https://support.google.com/analytics/answer/7586738}.
\newblock


\bibitem[Grover and Leskovec(2016)]%
        {grover2016node2vec}
\bibfield{author}{\bibinfo{person}{Aditya Grover} {and} \bibinfo{person}{Jure Leskovec}.} \bibinfo{year}{2016}\natexlab{}.
\newblock \showarticletitle{node2vec: Scalable feature learning for networks}. In \bibinfo{booktitle}{\emph{Proceedings of the 22nd ACM SIGKDD international conference on Knowledge discovery and data mining}}. \bibinfo{pages}{855--864}.
\newblock


\bibitem[Hadi et~al\mbox{.}(2023)]%
        {hadi2023survey}
\bibfield{author}{\bibinfo{person}{Muhammad~Usman Hadi}, \bibinfo{person}{qasem~al tashi}, \bibinfo{person}{Rizwan Qureshi}, \bibinfo{person}{Abbas Shah}, \bibinfo{person}{amgad muneer}, \bibinfo{person}{Muhammad Irfan}, \bibinfo{person}{Anas Zafar}, \bibinfo{person}{Muhammad~Bilal Shaikh}, \bibinfo{person}{Naveed Akhtar}, \bibinfo{person}{Jia Wu}, {and} \bibinfo{person}{Seyedali Mirjalili}.} \bibinfo{year}{2023}\natexlab{}.
\newblock \showarticletitle{{A Survey on Large Language Models: Applications, Challenges, Limitations, and Practical Usage}}.
\newblock  (\bibinfo{year}{2023}).
\newblock
\urldef\tempurl%
\url{https://doi.org/10.36227/techrxiv.23589741.v1}
\showDOI{\tempurl}


\bibitem[He et~al\mbox{.}(2023)]%
        {UBMforRec}
\bibfield{author}{\bibinfo{person}{Zhicheng He}, \bibinfo{person}{Weiwen Liu}, \bibinfo{person}{Wei Guo}, \bibinfo{person}{Jiarui Qin}, \bibinfo{person}{Yingxue Zhang}, \bibinfo{person}{Yaochen Hu}, {and} \bibinfo{person}{Ruiming Tang}.} \bibinfo{year}{2023}\natexlab{}.
\newblock \showarticletitle{A Survey on User Behavior Modeling in Recommender Systems}. In \bibinfo{booktitle}{\emph{the 32nd International Joint Conference on Artificial Intelligence}}. \bibinfo{pages}{6656--6664}.
\newblock


\bibitem[Ke et~al\mbox{.}(2017)]%
        {GuolinLightGBM}
\bibfield{author}{\bibinfo{person}{Guolin Ke}, \bibinfo{person}{Qi Meng}, \bibinfo{person}{Thomas Finley}, \bibinfo{person}{Taifeng Wang}, \bibinfo{person}{Wei Chen}, \bibinfo{person}{Weidong Ma}, \bibinfo{person}{Qiwei Ye}, {and} \bibinfo{person}{Tie-Yan Liu}.} \bibinfo{year}{2017}\natexlab{}.
\newblock \showarticletitle{LightGBM: a highly efficient gradient boosting decision tree}. In \bibinfo{booktitle}{\emph{Proceedings of the 31st International Conference on Neural Information Processing Systems}}. \bibinfo{pages}{3149–3157}.
\newblock
\showISBNx{9781510860964}


\bibitem[Keraghel et~al\mbox{.}(2024)]%
        {keraghel2024beyond}
\bibfield{author}{\bibinfo{person}{Imed Keraghel}, \bibinfo{person}{Stanislas Morbieu}, {and} \bibinfo{person}{Mohamed Nadif}.} \bibinfo{year}{2024}\natexlab{}.
\newblock \showarticletitle{Beyond words: a comparative analysis of LLM embeddings for effective clustering}. In \bibinfo{booktitle}{\emph{International Symposium on Intelligent Data Analysis}}. Springer, \bibinfo{pages}{205--216}.
\newblock


\bibitem[Kohavi et~al\mbox{.}(2020)]%
        {kohavi2020trustworthy}
\bibfield{author}{\bibinfo{person}{Ron Kohavi}, \bibinfo{person}{Diane Tang}, {and} \bibinfo{person}{Ya Xu}.} \bibinfo{year}{2020}\natexlab{}.
\newblock \bibinfo{booktitle}{\emph{Trustworthy online controlled experiments: A practical guide to a/b testing}}.
\newblock \bibinfo{publisher}{Cambridge University Press}.
\newblock


\bibitem[Kreinovich et~al\mbox{.}(2014)]%
        {kreinovich2014estimate}
\bibfield{author}{\bibinfo{person}{Vladik Kreinovich}, \bibinfo{person}{Hung~T. Nguyen}, {and} \bibinfo{person}{Rujira Ouncharoen}.} \bibinfo{year}{2014}\natexlab{}.
\newblock \showarticletitle{How to Estimate Forecasting Quality: A System- Motivated Derivation of Symmetric Mean Absolute Percentage Error (SMAPE) and Other Similar Characteristics}.
\newblock
\urldef\tempurl%
\url{https://api.semanticscholar.org/CorpusID:17589608}
\showURL{%
\tempurl}


\bibitem[Kumar et~al\mbox{.}(2020)]%
        {kumar2020link}
\bibfield{author}{\bibinfo{person}{Ajay Kumar}, \bibinfo{person}{Shashank~Sheshar Singh}, \bibinfo{person}{Kuldeep Singh}, {and} \bibinfo{person}{Bhaskar Biswas}.} \bibinfo{year}{2020}\natexlab{}.
\newblock \showarticletitle{Link prediction techniques, applications, and performance: A survey}.
\newblock \bibinfo{journal}{\emph{Physica A: Statistical Mechanics and its Applications}}  \bibinfo{volume}{553} (\bibinfo{year}{2020}), \bibinfo{pages}{124289}.
\newblock


\bibitem[Li et~al\mbox{.}(2024)]%
        {li2024condensed}
\bibfield{author}{\bibinfo{person}{Mingchen Li}, \bibinfo{person}{Chen Ling}, \bibinfo{person}{Rui Zhang}, {and} \bibinfo{person}{Liang Zhao}.} \bibinfo{year}{2024}\natexlab{}.
\newblock \showarticletitle{A Condensed Transition Graph Framework for Zero-shot Link Prediction with Large Language Models}.
\newblock \bibinfo{journal}{\emph{arXiv preprint arXiv:2402.10779}} (\bibinfo{year}{2024}).
\newblock


\bibitem[Li and Chen(2009)]%
        {li2009recommendation}
\bibfield{author}{\bibinfo{person}{Xin Li} {and} \bibinfo{person}{Hsinchun Chen}.} \bibinfo{year}{2009}\natexlab{}.
\newblock \showarticletitle{Recommendation as link prediction: a graph kernel-based machine learning approach}. In \bibinfo{booktitle}{\emph{Proceedings of the 9th ACM/IEEE-CS joint conference on Digital libraries}}. \bibinfo{pages}{213--216}.
\newblock


\bibitem[Microsoft Corporation(2024a)]%
        {AzureOpenAIService}
Microsoft Corporation \bibinfo{year}{2024}\natexlab{a}.
\newblock \bibinfo{booktitle}{\emph{Azure OpenAI Service}}.
\newblock Microsoft Corporation, \url{https://azure.microsoft.com/en-us/products/ai-services/openai-service/}.
\newblock
\newblock
\shownote{Accessed on May 29, 2024}.


\bibitem[Microsoft Corporation(2024b)]%
        {AzureOpenAIGPT3.5}
Microsoft Corporation \bibinfo{year}{2024}\natexlab{b}.
\newblock \bibinfo{booktitle}{\emph{Azure OpenAI Service Embedding Models} (\bibinfo{edition}{gpt-35-turbo (0125)} ed.)}.
\newblock Microsoft Corporation, \url{https://learn.microsoft.com/en-us/azure/ai-services/openai/concepts/models}.
\newblock


\bibitem[Microsoft Corporation(2024c)]%
        {AzureOpenAIGPT4}
Microsoft Corporation \bibinfo{year}{2024}\natexlab{c}.
\newblock \bibinfo{booktitle}{\emph{Azure OpenAI Service Embedding Models} (\bibinfo{edition}{gpt-4 (0125-preview)} ed.)}.
\newblock Microsoft Corporation, \url{https://learn.microsoft.com/en-us/azure/ai-services/openai/concepts/models}.
\newblock


\bibitem[Microsoft Corporation(2024d)]%
        {AzureOpenAIEmbeddingSmall}
Microsoft Corporation \bibinfo{year}{2024}\natexlab{d}.
\newblock \bibinfo{booktitle}{\emph{Azure OpenAI Service Embedding Models} (\bibinfo{edition}{text-embedding-3-small} ed.)}.
\newblock Microsoft Corporation, \url{https://learn.microsoft.com/en-us/azure/ai-services/openai/concepts/models#embeddings-models}.
\newblock


\bibitem[Microsoft Corporation(2024e)]%
        {AzureOpenAIEmbeddingLarge}
Microsoft Corporation \bibinfo{year}{2024}\natexlab{e}.
\newblock \bibinfo{booktitle}{\emph{Azure OpenAI Service Embedding Models} (\bibinfo{edition}{text-embedding-3-large} ed.)}.
\newblock Microsoft Corporation, \url{https://learn.microsoft.com/en-us/azure/ai-services/openai/concepts/models#embeddings-models}.
\newblock


\bibitem[Pan et~al\mbox{.}(2018)]%
        {pan2018adversarially}
\bibfield{author}{\bibinfo{person}{Shirui Pan}, \bibinfo{person}{Ruiqi Hu}, \bibinfo{person}{Guodong Long}, \bibinfo{person}{Jing Jiang}, \bibinfo{person}{Lina Yao}, {and} \bibinfo{person}{Chengqi Zhang}.} \bibinfo{year}{2018}\natexlab{}.
\newblock \showarticletitle{Adversarially regularized graph autoencoder for graph embedding}.
\newblock \bibinfo{journal}{\emph{arXiv preprint arXiv:1802.04407}} (\bibinfo{year}{2018}).
\newblock


\bibitem[Panagiotakis and Papadakis(2022)]%
        {panagiotakis2022session}
\bibfield{author}{\bibinfo{person}{Costas Panagiotakis} {and} \bibinfo{person}{Harris Papadakis}.} \bibinfo{year}{2022}\natexlab{}.
\newblock \showarticletitle{Session-based recommendation by combining probabilistic models and LSTM}.
\newblock In \bibinfo{booktitle}{\emph{Proceedings of the Recommender Systems Challenge 2022}}. \bibinfo{pages}{39--44}.
\newblock


\bibitem[Pavao et~al\mbox{.}(2016)]%
        {CIKM2016EComm}
\bibfield{author}{\bibinfo{person}{Adrien Pavao}, \bibinfo{person}{Isabelle Guyon}, \bibinfo{person}{Anne-Catherine Letournel}, \bibinfo{person}{Dinh-Tuan Tran}, \bibinfo{person}{Xavier Baro}, \bibinfo{person}{Hugo~Jair Escalante}, \bibinfo{person}{Sergio Escalera}, \bibinfo{person}{Tyler Thomas}, {and} \bibinfo{person}{Zhen Xu}.} \bibinfo{year}{2016}\natexlab{}.
\newblock \bibinfo{booktitle}{\emph{CIKM Cup 2016 Track 2: Personalized E-Commerce Search Challenge}}.
\newblock CodaLab, \url{https://competitions.codalab.org/competitions/11161}.
\newblock
\newblock
\shownote{Accessed on April 28, 2024}.


\bibitem[Pavao et~al\mbox{.}(2023)]%
        {codalab_competitions_JMLR}
\bibfield{author}{\bibinfo{person}{Adrien Pavao}, \bibinfo{person}{Isabelle Guyon}, \bibinfo{person}{Anne-Catherine Letournel}, \bibinfo{person}{Dinh-Tuan Tran}, \bibinfo{person}{Xavier Baro}, \bibinfo{person}{Hugo~Jair Escalante}, \bibinfo{person}{Sergio Escalera}, \bibinfo{person}{Tyler Thomas}, {and} \bibinfo{person}{Zhen Xu}.} \bibinfo{year}{2023}\natexlab{}.
\newblock \showarticletitle{CodaLab Competitions: An Open Source Platform to Organize Scientific Challenges}.
\newblock \bibinfo{journal}{\emph{Journal of Machine Learning Research}} \bibinfo{volume}{24}, \bibinfo{number}{198} (\bibinfo{year}{2023}), \bibinfo{pages}{1--6}.
\newblock
\urldef\tempurl%
\url{http://jmlr.org/papers/v24/21-1436.html}
\showURL{%
\tempurl}


\bibitem[REES46 Inc.(2024)]%
        {REES46}
REES46 Inc. \bibinfo{year}{2024}\natexlab{}.
\newblock \bibinfo{booktitle}{\emph{REES46 for eCommerce}}.
\newblock REES46 Inc., \url{https://rees46.com/}.
\newblock
\newblock
\shownote{Accessed on April 28, 2024}.


\bibitem[Reisenbichler et~al\mbox{.}(2022)]%
        {NLPForMarketing}
\bibfield{author}{\bibinfo{person}{Martin Reisenbichler}, \bibinfo{person}{Thomas Reutterer}, \bibinfo{person}{David~A Schweidel}, {and} \bibinfo{person}{Daniel Dan}.} \bibinfo{year}{2022}\natexlab{}.
\newblock \showarticletitle{Frontiers: Supporting content marketing with natural language generation}.
\newblock \bibinfo{journal}{\emph{Marketing Science}} \bibinfo{volume}{41}, \bibinfo{number}{3} (\bibinfo{year}{2022}), \bibinfo{pages}{441--452}.
\newblock


\bibitem[Shu et~al\mbox{.}(2024)]%
        {shu2024knowledge}
\bibfield{author}{\bibinfo{person}{Dong Shu}, \bibinfo{person}{Tianle Chen}, \bibinfo{person}{Mingyu Jin}, \bibinfo{person}{Yiting Zhang}, \bibinfo{person}{Mengnan Du}, {and} \bibinfo{person}{Yongfeng Zhang}.} \bibinfo{year}{2024}\natexlab{}.
\newblock \showarticletitle{Knowledge Graph Large Language Model (KG-LLM) for Link Prediction}.
\newblock \bibinfo{journal}{\emph{arXiv preprint arXiv:2403.07311}} (\bibinfo{year}{2024}).
\newblock


\bibitem[Trouillon et~al\mbox{.}(2016)]%
        {trouillon2016complex}
\bibfield{author}{\bibinfo{person}{Th{\'e}o Trouillon}, \bibinfo{person}{Johannes Welbl}, \bibinfo{person}{Sebastian Riedel}, \bibinfo{person}{{\'E}ric Gaussier}, {and} \bibinfo{person}{Guillaume Bouchard}.} \bibinfo{year}{2016}\natexlab{}.
\newblock \showarticletitle{Complex embeddings for simple link prediction}. In \bibinfo{booktitle}{\emph{International conference on machine learning}}. \bibinfo{pages}{2071--2080}.
\newblock


\bibitem[Wang et~al\mbox{.}(2024)]%
        {wang2024exploring}
\bibfield{author}{\bibinfo{person}{Yiqi Wang}, \bibinfo{person}{Wentao Chen}, \bibinfo{person}{Xiaotian Han}, \bibinfo{person}{Xudong Lin}, \bibinfo{person}{Haiteng Zhao}, \bibinfo{person}{Yongfei Liu}, \bibinfo{person}{Bohan Zhai}, \bibinfo{person}{Jianbo Yuan}, \bibinfo{person}{Quanzeng You}, {and} \bibinfo{person}{Hongxia Yang}.} \bibinfo{year}{2024}\natexlab{}.
\newblock \showarticletitle{Exploring the reasoning abilities of multimodal large language models (mllms): A comprehensive survey on emerging trends in multimodal reasoning}.
\newblock \bibinfo{journal}{\emph{arXiv preprint arXiv:2401.06805}} (\bibinfo{year}{2024}).
\newblock


\bibitem[Wu et~al\mbox{.}(2019)]%
        {wu2019session}
\bibfield{author}{\bibinfo{person}{Shu Wu}, \bibinfo{person}{Yuyuan Tang}, \bibinfo{person}{Yanqiao Zhu}, \bibinfo{person}{Liang Wang}, \bibinfo{person}{Xing Xie}, {and} \bibinfo{person}{Tieniu Tan}.} \bibinfo{year}{2019}\natexlab{}.
\newblock \showarticletitle{Session-based recommendation with graph neural networks}. In \bibinfo{booktitle}{\emph{Proceedings of the AAAI conference on artificial intelligence}}, Vol.~\bibinfo{volume}{33}. \bibinfo{pages}{346--353}.
\newblock


\bibitem[Zhang et~al\mbox{.}(2020)]%
        {ZHANG202040}
\bibfield{author}{\bibinfo{person}{Fan Zhang}, \bibinfo{person}{Yiqun Liu}, \bibinfo{person}{Jiaxin Mao}, \bibinfo{person}{Min Zhang}, {and} \bibinfo{person}{Shaoping Ma}.} \bibinfo{year}{2020}\natexlab{}.
\newblock \showarticletitle{User behavior modeling for Web search evaluation}.
\newblock \bibinfo{journal}{\emph{AI Open}}  \bibinfo{volume}{1} (\bibinfo{year}{2020}), \bibinfo{pages}{40--56}.
\newblock
\showISSN{2666-6510}
\urldef\tempurl%
\url{https://doi.org/10.1016/j.aiopen.2021.02.003}
\showDOI{\tempurl}


\end{thebibliography}

\end{document}